\documentclass[journal, twoside]{IEEEtran}
\usepackage{xspace}
\newcommand*{\eg}{e.g.\@\xspace}
\newcommand*{\ie}{i.e.\@\xspace}

\ifCLASSINFOpdf
\else
\fi
\usepackage{graphicx}
\graphicspath{{images/}}

\usepackage{amsfonts}

\usepackage{bm}

\usepackage{mathtools}

\usepackage{amsmath} 

\usepackage{textcomp, gensymb}

\usepackage{amssymb}

\usepackage{nicefrac}

\makeatletter
\newcommand*\bigcdot{\mathpalette\bigcdot@{.5}}
\newcommand*\bigcdot@[2]{\mathbin{\vcenter{\hbox{\scalebox{#2}{$\m@th#1\bullet$}}}}}
\makeatother

\usepackage[caption=false,font=normalsize,labelfont=sf,textfont=sf]{subfig}

\usepackage{hyperref}

\usepackage[prependcaption,colorinlistoftodos]{todonotes}

\renewcommand{\baselinestretch}{0.99}

\mathchardef\mhyphen="2D

\usepackage{amsthm}
\newtheorem{remark}{Remark}
\newtheorem{proposition}{Proposition}

\newtheorem{hypothesis}{Hypothesis}

\usepackage[ruled,vlined]{algorithm2e}

\begin{document}
%
\title{A New Human-Likeness and Comfort Index for Robot Movements Along Prescribed Paths}
%
%
%

\author{Rosanna Coccaro, Enrico Ferrentino, Antonio Parziale, Angelo Marcelli and Pasquale Chiacchio%
\thanks{This research was conducted in the frame of the Department of Excellence 2023/2027 Project and partially funded by \textit{Istituto Nazionale della Previdenza Sociale} (INPS).}
\thanks{The authors are with the Department of Computer Engineering, Electrical Engineering and Applied Mathematics (DIEM), University of Salerno, 84084 Fisciano, Italy
        (e-mail: {\tt\footnotesize rcoccaro@unisa.it}; {\tt\footnotesize eferrentino@unisa.it}; {\tt\footnotesize anparziale@unisa.it}; {\tt\footnotesize amarcelli@unisa.it}; {\tt\footnotesize pchiacchio@unisa.it})}
\thanks{
This article has supplementary material provided by the authors available at \url{https://dx.doi.org/10.21227/amq5-qp90}. The source code is also available at \url{https://github.com/unisa-acg/trajectory_planning_matlab/tree/v01r00}.

©2026 IEEE. Personal use of this material is permitted. Permission
from IEEE must be obtained for all other uses, in any current or future
media, including reprinting/republishing this material for advertising or
promotional purposes, creating new collective works, for resale or
redistribution to servers or lists, or reuse of any copyrighted component
of this work in other works.}%
}
%
%

\markboth{Accepted version, published at 10.1109/TCYB.2026.3707010 – 2026 IEEE Transactions on Cybernetics}
{Accepted version, published at 10.1109/TCYB.2026.3707010 – 2026 IEEE Transactions on Cybernetics} 

%



\maketitle

\begin{abstract}
As human-robot interaction rapidly spreads in numerous fields, the subject of robot acceptance gains increasing importance. Visual similarity to the human body, as occurs for humanoids, is generally not enough to ensure acceptance in physical interaction, as acceptance directly links to comfort and ergonomics, which are measured in terms of the quality of the robot movement perceived by the human. This paper discusses the connection between comfort and similarity of the robot movement to the human one. By considering the kinematic characterization of human movement, this paper focuses on the time laws of such movements, wherein the end-effector path is prescribed. Based on the lognormality principle for modeling human movements, a human-likeness index is defined and used to provide an a priori characterization of trajectories. Such an index can be used to evaluate the performance of trajectory generation algorithms in producing human-like movements before they are actually executed. For validation purposes, 68 subjects are required to judge their comfort. The results of three experimental campaigns involving a physical interaction with a robot demonstrate a globally consistent trend between the preference in terms of perceived comfort and the distribution of the suggested human-likeness index.
\end{abstract}


%
\IEEEpeerreviewmaketitle

\section{Introduction} \label{sec:intro}

\IEEEPARstart{S}{ince} the emergence of Human-Robot Interaction (HRI) as a field, several studies have concerned the ``social'' dimension of robots in industry, particularly, in relation to cognitive and perceptual workload for robot operators \cite{monizRobotsWorkingHumans2016}, that can undermine acceptance. 
A strategy to tackle user robot acceptance goes through the design of robots resembling humans in both their physical aspect and their behavior, as described in \cite{hutchison_anthropomorphism_2012}.
To this extent, recent research has led to the development of brain-inspired intelligent robots that are capable of emulating humans and animals in both their external structures and internal mechanisms. This is achieved through the integration of visual cognition, decision-making, motion control, and musculoskeletal systems \cite{qiaoSurveyBrainInspiredIntelligent2022, ParzialeSurvey2024, Su_2025}.
A fundamental premise is that humans tend to favor engaging with machines (and robots) \cite{yuan_dynamic_2025} in a manner that resembles their interactions with other humans \cite{natarajan_effects_2020, nie_2024}. This is sustained by the concept of \emph{motor contagion}, which posits that an individual's motor actions are involuntarily influenced by observing and experiencing others’ movements. The degree of motor contagion is influenced by the human likeness of a robot: the greater the resemblance between robot and human movements and appearance, the stronger the motor contagion effect. Movement kinematics, such as the velocity profile and the trajectory of the limb, might also vary the degree of motor resonance evoked in the observer. This is demonstrated in \cite{bisio2014}, where participants were asked to reproduce an action after observing either a human or a robot performing it. Motor contagion occurred in both cases, except when the robot violated the biological laws of motion. This supports the idea that the observer relies on recognizing elements of its own motor repertoire in the observed model. Likewise, \cite{baddoura_human_2014} suggests that exhibiting a human-like movement of the robot results in an easier prediction by the human counterpart. On the other hand, fast movements have been observed to increase anxiety and risk perception among human co-workers, who, consequently, exhibit unpleasant and inefficient behavior. Therefore, the evidence suggests that time parametrization is crucial and worth of investigation when designing motion planning processes for human-robot interaction \cite{gulletta_2020}. Indeed, recent works explicitly propose planning for human-like velocity profiles to achieve natural motion \cite{2025_gaebert}.

Based on the current literature, our study formulates  the following
\begin{hypothesis} \label{hp:hl_comfort}
    If $A$ and $B$ correspond to, respectively, ``a movement is similar to a human one'' and ``a movement is comfortable'', then:
    \begin{equation}
        A \iff B.
    \end{equation}
\end{hypothesis}

In the above statement, the term \emph{movement} is used to denote the time law of the movement, as the geometrical component, i.e.\ the path, is notably task-dependent. Among the many facets of human-robot interaction, we are particularly interested in, and therefore this study focuses on, evaluating the comfort of a human interacting solely with the robot's end-effector, as in collaborative assembly and transportation tasks. Therefore, our hypothesis states that a human will perceive the robot's end-effector movement as comfortable as that of a human teammate if and only if the former exhibits the same time law as the latter. From now on, for the sake of brevity, we will use the term \emph{movement} as specified above.

Thus, in this paper, we aim to evaluate the comfort by measuring the similarity of the robot movement with the corresponding human one, which we term \emph{human-likeness}, so addressing a notable gap in the literature that consists in the absence of a quantitative index for such a similarity \cite{gulletta_2020, kulic_anthropomorphic_2016}. Moreover, as it follows from above, such a measure can be employed in trajectory planning to obtain a comfortable-by-design movement, avoiding the burden of the trial-and-error approach adopted by data-driven strategies. The definition of an objective measure accounting for the specific characteristics of the human movement represents a significant advancement in the field. Therefore, in what follows, through a large experimental campaign of collaborative human-robot interaction sessions focused on comfort, we validate our index and, in turn, we also confirm the validity of Hypothesis \ref{hp:hl_comfort} in our data.

\subsection{Related works} \label{sec:related_works}

Two main branches can be identified in the literature dealing with human likeness in robotics: one regarding human-like robotic movements, and another concerning means for comfort evaluation. For the latter, one of two approaches is typically adopted: one focuses on the human perspective and aims to measure how humans behave or perceive the robot, as in \cite{marchesi_i_2021, ghiglino_at_2020, ghiglino_can_2020, yan_2025}, the other is centered on the robot's actual movement and is usually based on metrics such as repeatability \cite{koskinopoulou2016}, trajectory smoothness \cite{chang2005}, or the two-thirds power law \cite{demomi2016}. Recent advancements within this second domain have expanded toward task-specific comfort metrics combining smoothness, velocity, and proxemics considerations \cite{maheshwari_2025}, as well as ergonomic indicators derived from human posture and task execution \cite{ribeiro_2026, proia_2025}. We thus focus on works that bridge these two domains, with the aim of studying comfort through the similarity to the human movement.

An example is found in \cite{heAnthropomorphicReachingMovement2022}, in which the authors propose a method for generating human-like reaching movements in robots, that produces smooth trajectories and allows for natural obstacle avoidance. The method is validated through experiments with an anthropomorphic upper limb robot. An indicator of human likeness is defined, consisting of the difference between the robot trajectory and the human reference one. This indicator, that is, in fact, a prevalent way to assess human likeness, allows having a quantitative measure of such similarity, but it cannot be of general use, as it can be exclusively computed after the trajectory has been executed. Furthermore, the results demonstrate that the degree of human likeness is dependent on the specific human subject, leading to an additional lack of generality. Similarly, \cite{2026_tamura} introduces objective engineering metrics computed directly from the robot's motion data and demonstrates that such metrics correlate with human perception and can be incorporated into control optimization loops. However, despite their methodological differences, both approaches assess human-likeness from executed motion and remain tied to specific trajectories, control parameters, and hardware characteristics. This limitation further motivates the importance of capturing common intrinsic features of human kinetics rather than mapping human trajectories to robot kinematics, which often demands sophisticated domain adaptation architectures to handle cross-subject variability \cite{yin_liu_2025}.

In \cite{ryuHumanoidPathPlanning2013}, the authors propose a path planner for a humanoid robot, aiming to enhance its performance in terms of HRI. The proposed method employs a time index, with the objective of generating a path that is perceived as natural by humans. A survey is conducted on humans' preferences. Participants are shown a set of video clips in which a simulated robot traces two different paths. For each video, they are asked to express whether one of the two planned paths feels more natural or if no difference is discerned. This research diverges from our approach as we pursue human likeness by examining time laws, whereas the authors investigate the planned path. Moreover, the authors do not propose a quantitative human-likeness measure, as surveys are employed in path evaluation.

Similarly, in \cite{quintana_uniform_2022}, experiments are conducted to analyze, through surveys, the human perception of both artificial and human movements. Human movements are generated using the sigma-lognormal model \cite{plamondon_generation_1998}, while the artificial movements have a trapezoidal velocity profile \cite{siciliano_motion_2016}. The robot is programmed to perform triangular and square shapes in its free space, and, in one of the proposed experimental setups, each participant has to touch the robot's end-effector with their index finger to track its movements.
Participants are asked to rate the comfort and friendliness of each movement on a scale. The study reveals that humans do not display a clear preference for either human or robotic movement when observing the motion or interacting with a robot by just tracking its end-effector.
In this regard, some significant considerations arise concerning the proposed human movement. Despite its intention to resemble natural motion, it may not achieve sufficient human likeness, and this aspect could potentially contribute to the observed results. Furthermore, it is worth noting that the interaction experiment solely involves tracking rather than an active physical engagement with the robot. This methodological distinction may influence the human perception of movement, introducing potential variations in comfort evaluation.

\subsection{Problem statement} \label{sec:problem_statement}

The issue addressed in this study is that, despite increasing interest in exploring the concept of comfort in HRI, currently there is no global quantitative index of comfort that can be computed without the need to experience the movement itself \cite{gulletta_2020}. The available indices fall into two main categories: qualitative, a posteriori measures, \eg questionnaires, that aim to assess comfort after interaction has occurred \cite{dragan2013, howell_effects_2023}; quantitative measures, that primarily focus on local motion characteristics. In this latter case, a key challenge lies in integrating multiple quantitative measures related to different aspects of the movement, \eg human-robot proximity \cite{yan_comfort_factors_2022}, jerk-based measures \cite{hogan_sensitivity_2009} and legibility \cite{dragan2013}.


It is also worth noting that measures such as jerk or power-spectral-density-based ones are not direct indicators of comfort. Rather, they may only capture necessary but not sufficient conditions for comfort, as comfortable interactions are characterized by smooth movements \cite{rozlivek_2025}.

Finally, since in most application scenarios it is desirable for the planning to be independent of prior experience, it is crucial to have an a priori comfort index. Such an index would enable the planning for human-like trajectories without relying on post-hoc assessments.

\subsection{Statement of contribution}

As discussed above, the concept of motor contagion provides the foundation for formulating Hypothesis \ref{hp:hl_comfort}. The main contribution of this paper is to provide the empirical evidence validating it. Along this line, we introduce a comfort index that enables the evaluation of a trajectory's comfort before making an experience of it, addressing the problem stated in Section \ref{sec:problem_statement}. On the basis of Hypothesis \ref{hp:hl_comfort}, the comfort index is designed as a human likeness index. Therefore, as will be discussed in Section \ref{sec:sigma_lognormal}, relying on the sigma-lognormal model of velocity \cite{oreilly_development_2009}, it evaluates comfort indirectly, through the level of resemblance of the robot behavior to the human one. Notably, this index addresses a gap in the existing literature, namely the absence of a quantitative mean of assessment by which to obtain objective information concerning the human likeness of robotic trajectories \cite{gulletta_2020}, prior to their execution. Unlike other proposed indices, our index is not task-related, thus dependent on the performance of single interactions, but it captures global characteristics of comfort.
	
As an additional contribution, we provide a planning technique that, within certain limits, can generate human-like movements for arbitrary paths. We employ this technique in our validation procedure.

Lastly, our approach differs from others in that it decouples the geometric path assigned to the robot end-effector and its time law. We associate comfort to the latter, so that, in principle, any movement performed by the robot can be human-like (and comfortable), regardless of its geometry. 
This would extend the applicability of human-like movements to those industrial and medical scenarios where robots are required to accurately track specific paths, while possibly interacting with humans.

\subsection{Outline}

This article is organized as follows. In Section \ref{sec:materials}, we provide an overview of the essential concepts on which the proposed human-likeness index is based, alongside the definition of such index. We recall some techniques used for time parametrization of robotic paths, as these will be employed in Section \ref{sec:index_validation_setup}, throughout the index validation procedure. Section \ref{sec:index_validation_setup} portrays the methodology used for the validation of the human-likeness index. A first numerical validation of the index is presented, leading to the introduction of interaction experiments. The experimental setup is presented in Section \ref{sec:experimental_setup} and experimental results are analyzed and discussed in Section \ref{sec:results_analysis}, with a focus on describing the advantages and disadvantages of our index and methodology. Finally, we draw conclusions and highlight future developments in Section \ref{sec:conclusion}.

\section{Materials and methods} \label{sec:materials}

Given that the concept of ``behavior'' covers a range of different aspects, there is to specify that this paper is concerned with the kinematic characteristics of movement. Among these, we focus on the time law that underlies movement, assuming that the geometrical features of said motion are already defined. Our focus therein lies in investigating whether the human likeness of robot trajectories can be assessed only on their time laws, regardless of their geometric paths, which would enable preserving the level of accuracy required from a robot during the execution of a specific task. This can be required in tasks such as collaborative welding \cite{antonelli_qualification_2016} and rehabilitation \cite{trajectory_learning_2023}, where a path is prescribed, and the human coworker is expected to comply with the robot as it executes the movement. While we recognize that the internal robot movement may appear less visually pleasant when the robot lacks anthropomorphic physical characteristics, analyzing the impact of robot appearance or internal movements on the user's perceived comfort is beyond the scope of this work. We further clarify that, when we refer to movement, we are concerned only with the end-effector motion, not the overall joint movements of the robot.

\subsection{Motor program and sigma-lognormal model} \label{sec:sigma_lognormal}

Hypothesis \ref{hp:hl_comfort} sets an identity between comfort and human likeness. But, when can a movement be said to be similar to a human one?

The ability of humans and animals to accomplish a complex motor action using different end-effectors is a phenomenon known as ``motor equivalence''. Studies on this phenomenon suggest that the central nervous system memorizes motor actions through an encoding independent of the motor commands required to move a specific end-effector~\cite{wing2000motor}. This abstract representation of an action is known as \emph{motor program}, which has been defined as ``\textit{a central representation of a sequence of motor actions}''~\cite{summersCurrentStatusMotor2009}. The concept of motor program originated in the work of Lashley~\cite{lashleyProblemSerialOrder1951}, where the author conjectured the existence of a repertoire of ``units of action'' that can be preselected and activated by trigger commands, specifying the order in which the units of actions have to be executed.

At the beginning of the learning process, the central nervous system controls each unit of action separately, aiming to reach a 3-D target point in space that has been visually selected. During this initial stage, movements are executed in a stop-and-go modality, as each unit of action starts only after the previous one has finished. Such a strategy is slow and energetically expensive, but it allows for the learner to exploit the feedback information to correct errors and be able to track a trajectory that is more accurate in terms of motor production and more economical in terms of metabolic energy expenditure~\cite{sparrow1998metabolic} as the learning goes on. Once learning is consolidated, the sequence of target points is no longer considered as composed of isolated steps, but encoded as a single, continuous behavior, in which each unit of action is commanded and executed before the previous one terminates.

So, a motor program is an abstract representation of a complex movement, developed through repeated practice, aimed at accurately performing the movement while minimizing the consumption of metabolic energy, \ie it consists of the minimal set of units of action necessary to perform the movement it encodes.

We thus answer the opening question of this section by setting the following
\begin{proposition} \label{prop:hl_mp}
A sufficient condition for a movement to be human-like is for it to be the realization of a motor program.
\end{proposition}

Amidst the discussions and developments surrounding the motor program concept, the kinematic theory of rapid human movements \cite{plamondon_kinematic_1995} and its sigma-lognormal model \cite{oreilly_development_2009}
suggest that such ``units of action'', called \emph{strokes}, are lognormal impulse responses produced by a neuromuscular system activated by impulsive commands delivered by the central nervous system.

The lognormal function equation is
\begin{equation}
    \Lambda (t;t_0,\mu,\sigma) = \frac{1}{\sigma\sqrt{2\pi}(t-t_0)} \exp \left \{ -\frac{1}{2\sigma^2} [\ln(t-t_0)-\mu]^2\right \},
\end{equation}
where $t$ represents the time, $t_0$ is the time shift and $\mu$ and $\sigma$ are the mean and the standard deviation of the distribution on a logarithmic scale, respectively.

Let ${\bf x}_j(t) \in \mathbb{R}^3$ be the j-th stroke of a movement in the three-dimensional space. According to the mentioned models, it exhibits a lognormal velocity norm that results from the activation of either the agonist or the antagonist neuromuscular system recruited to perform the movement \cite{oreilly_development_2009}, i.e.
\begin{equation} \label{eq:stroke_vel_profile}
    \left \| \dot{\bf x}_j(t) \right \| = D_j\Lambda(t;t_j,\mu_j,\sigma_j),
\end{equation}
in which the dot notation indicates the time derivative. Therefore, $D_j$ is the amplitude of the command given for the generation of the neuromuscular component, $t_j$ is the time of the command's emission, $\mu_j$ and $\sigma_j$ represent the time delay and response time of the neuromuscular impulse response, respectively.

Within this framework, complex human movements can then be conceptualized as composed by a time superimposition of strokes, i.e.\ agonist and antagonist activation. Hence, the velocity profile $\dot{\bf x}_r$ of a complex human movement is modeled as a vectorial summation of $N$ lognormal elementary movements:
\begin{equation} \label{eq:complex_movement}
    \dot{\bf x}_r(t) =\textstyle\sum_{j=1}^{N}\dot{\bf x}_{j}(t).
\end{equation}
The effect of time superimposition of strokes is that target points are never reached and inferring the motor program by analyzing the executed trajectory is not trivial.
It is also important to note that the resulting movement is not determined just by the motor program, as additional contributions from the spinal cord allow fine adjustments during execution, meaning that some strokes originate from these adjustments rather than directly from the motor program \cite{plamondon_extracting_2020}.
In the literature, there are specific algorithms that, from a complex movement (measured/observed), perform stroke extraction and parameter estimation, i.e., computing $\dot{\bf x}_r(t)$ through an iterative process. 
These methods lead to trajectory reconstruction, both in the spatial and in the kinematic domain, by means of an optimization problem that aims at maximizing the \emph{Signal-to-Noise Ratio} (SNR) between the observed movement and its reconstruction $\dot{\bf x}_r(t)$, assessed in regard to both the geometric path and velocity profile. Given their inherent correlation, optimizing one aspect might result in a compromise with the other. Such algorithms can differ in terms of stroke definition and performance over the SNR measurement. A detailed discussion on strokes extraction and movement reconstruction is beyond the scope of this work, thus, for specific examples, we refer the interested reader to \cite{oreilly_development_2009, ferrerIDeLogIterativeDual2020, plamondon_extracting_2020}. We further support our choice of the sigma-lognormal model as reconstructions based on it are found to be difficult to distinguish from genuine human movements \cite{leiva_plamondon_2025}.


For this study, we choose the Robust Zero crossing (RX0) algorithm available in the ScriptStudio (SS) software \cite{oreilly_development_2009}, as it exhibits satisfactory performances on the reconstruction of both the geometric path and the velocity profile\footnote{According to \cite{ferrerIDeLogIterativeDual2020}, SS may struggle to recover long and complex velocity profiles. In such a case, start and stop (cuspidal) points shall be identified in the geometric path to break it into smaller segments, and then use SS on each segment. Alternatively, different sigma-lognormal reconstruction algorithms shall be considered for the computation of the HL index, \eg iDeLog. We remark that, while the latter approach may have an impact on the absolute value of the HL index (by the effect of possibly changing both $N$ and the SNR), it would not prevent the user from performing comparative analyses, as proposed in this paper.}. The latter is the factor that we aim to isolate and examine. This approach, which decouples geometry and kinematics, is justified by the fact that, in \eqref{eq:complex_movement}, $\dot{\bf x}_{j}$ is only constrained in its norm, according to \eqref{eq:stroke_vel_profile}, suggesting that, in principle, human-like profiles can be assigned to arbitrary geometrical paths and that, in other words, human likeness is independent of the geometrical features of the path.

Interestingly, we note that this velocity-centered approach is consistent with the recent literature on primitives-based planning and has remarkable applications going beyond comfort evaluation \cite{wang_2025}.

\subsection{Phase plane representation of time laws} \label{sec:phase_plane_analysis}

The phase plane analysis is a typical approach in robotic planning used to isolate time laws and analyze them regardless of the underlying geometrical path, which can be defined either in joint space or in task space. In the latter case, the path is ${\bf x}(\lambda) \in \mathbb{R}^m$, where $\lambda$ is the path parameter and, if the robot cannot stop or reverse its motion, it is a strictly monotonic function of time, \ie
\begin{equation}
\lambda(t): \mathbb{R} \rightarrow [0;L], \quad \dot{\lambda}=\frac{d\lambda}{dt} > 0 \; \quad  \forall t.
\end{equation}
\begin{remark}
\label{rem:lambda}
$L$ is chosen as the length of the path.
\end{remark}

The function $\lambda(t)$, or equivalently, its inverse $t(\lambda)$, represents the time law, and it has to be found as a result of the time parametrization process. The phase plane analysis deals with determining $\lambda(t)$ indirectly, through the definition of curves in the  $\lambda \mhyphen \dot{\lambda}$ plane, i.e.\ plotting the pseudo-velocity $\dot{\lambda}$ with respect to the path parameter $\lambda$. $\dot{\lambda}(\lambda)$ is termed \emph{Phase Plane Trajectory} (PPT), from which the time can be computed as
\begin{equation} \label{eq:time_law}
    t(\lambda) = \int_{0}^{\lambda} \frac{1}{\dot{\lambda}(l)}dl.
\end{equation}

Once a time law is available, it can be assigned to a path to obtain a \emph{trajectory}, i.e.\ ${\bf x}\big( \lambda(t) \big) = {\bf x}(t)$. Let ${\bf x}'(\lambda)=\frac{d {\bf x}(\lambda)}{d\lambda}$ and ${\bf x}''(\lambda) = \frac{d^2 {\bf x}(\lambda)}{d\lambda^2}$. By using the chain rule on $\dot{\bf x}\big( \lambda(t) \big)$ and $\ddot{\bf x}\big( \lambda(t) \big)$, we obtain the equations 
\begin{align}
\dot{\bf x} &= {\bf x}' \dot{\lambda} \label{eq:task_velocity}, \\
\ddot{\bf x} &= {\bf x}' \ddot{\lambda} + {\bf x}'' \dot{\lambda}^2, \label{eq:task_acceleration}
\end{align}
which link the task velocity and acceleration $\dot{\bf x}(t)$ and $\ddot{\bf x}(t)$ with their parametric velocity and acceleration ${\bf x}'(\lambda)$ and ${\bf x}''(\lambda)$. To this respect, we note that the pseudo-velocity $\dot{\lambda}$ can also be regarded as the magnitude of the tangential velocity $\left \| \dot{\bf x}(t) \right \|$ (when given with respect to time) or $\left \| \dot{\bf x}(\lambda) \right \|$ (when given with respect to the path parameter).

If Remark \ref{rem:lambda} holds, ${\bf x}'(\lambda)$ (also termed $\lambda$\emph{-velocity} or \emph{parametric velocity}) is the unit vector tangent to the path, and ${\bf x}''(\lambda)$ (also termed $\lambda$\emph{-acceleration} or \emph{parametric acceleration}) is a vector normal to the path, whose magnitude depends on the path curvature. Therefore, we have that \cite{zlajpah_implementation_1999}:
\begin{align}
\left \| {\bf x}' \right \| = 1 \label{eq:unit_norm}, \\
{\bf x}'^{T}{\bf x}'' = 0. \label{eq:orthogonality}
\end{align}

\subsection{Human-Likeness index definition}

When a trajectory ${\bf x}(t)$ is given (either through observation of human movements or by a planning algorithm) we can reconstruct its best sigma-lognormal approximation $\dot{\bf x}_r(t)$ according to \eqref{eq:complex_movement}, with one of the algorithms mentioned in Section \ref{sec:sigma_lognormal}. From this, we isolate the PPT and evaluate the performance of trajectory reconstruction, by computing the SNR on the time law only:
\begin{equation} \label{eq:snr_reconstruction}
    SNR = 10 \log \frac{\int_{0}^{1} \dot{\lambda}^2(s) \,ds}{\int_{0}^{1} (\dot{\lambda}(s) - \dot{\lambda}_{r}(s))^2 \,ds},
\end{equation}
where $\dot{\lambda}(\lambda)$ is the PPT associated to $\dot{\bf x}(t)$ and $\dot{\lambda}_r(\lambda)$ is that associated to the reconstructed movement $\dot{\bf x}_r(t)$. Note that the integration is performed over the path, normalized with respect to its length $L$. Therefore, $s \in [0,1]$ represents the achieved path fraction.

A large SNR in \eqref{eq:snr_reconstruction} attests to a high degree of similarity between the given trajectory ${\bf x}(t)$ and its sigma-lognormal reconstruction ${\bf x}_r(t)$. Therefore, one may conclude that the SNR alone is a human-likeness index. However, any movement can be accurately approximated with an arbitrarily large number of strokes $N$, while human movements, being the realization of well-developed motor programs, are composed of a minimal set of strokes (see Proposition \ref{prop:hl_mp} and related discussion). Therefore, reconstructions characterized by many strokes must be penalized in virtue of representing the typical human behavior of optimizing metabolic energy consumption. Hence, we define the human-likeness index $HL$ as
\begin{equation} \label{eq:human-likeness}
    HL = SNR~\frac{\bar{N}}{N},
\end{equation}
where $\bar{N}$ is the minimum amount of strokes necessary to encode the motor program of the assigned path, so that $N \geq \bar{N}$ always holds, and $0 < \bar{N}/N \leq 1$ is a measure of the movement's simplicity.

Since the $HL$ index is a measure of similarity to a motor program, we can elaborate on Proposition \ref{prop:hl_mp} to state
\begin{proposition} \label{prop:max_hl_implies_human_likeness}
For a given movement, if $HL$ is maximized, the movement is human-like.
\end{proposition}

\begin{remark} \label{rem:sufficient_condition}
    Considering that the $HL$ index is an index of similarity to a sigma-lognormal motor program and that Proposition \ref{prop:max_hl_implies_human_likeness} only provides a sufficient condition, we should expect that movements exist that are human-like, but whose $HL$ index is low. These shall comprise the movements including additional strokes generated by the spinal cord signals based on proprioceptive information.
\end{remark}

\subsection{Time parametrization of robotic paths} \label{sec:time_para_robotic_paths}

Before introducing the experimental setup to validate \eqref{eq:human-likeness}, it is worth recalling some basic planning concepts in the phase plane. Indeed, we will use this technique to generate ``surrogates'' of human movements in the validation process. As we said in Section \ref{sec:phase_plane_analysis}, planning in the phase plane amounts to defining the PPT $\dot{\lambda}(\lambda)$. In this work, we adopt, for each time parametrization, one of the following methods.

\subsubsection{Constant velocity profile} \label{sec:uniform_theory}

since $\dot{\lambda}(\lambda)$ is the tangential velocity's norm, assigning a constant velocity profile in the phase plane is as simple as setting
\begin{equation} \label{eq:timestamps}
    \dot{\lambda}(\lambda) = k,
\end{equation}
where $k \in \mathbb{R}^+$ is a constant.

\subsubsection{Time-Optimal Trajectory Planning (TOTP)} \label{sec:totp_theory}

in TOTP \cite{slotine_improving_1989}, the objective is to find $\dot{\lambda}(\lambda)$ such as to minimize the trajectory tracking time, subject to application-dependent constraints.

Since our objective is that of emulating human movements, and robots have actuation capacity beyond human abilities, we disregard physical robot limits, usually given in the joint space, e.g., maximum motor velocity, maximum actuation torque, and only consider task space constraints in terms of acceleration and velocity limits.
In the definition of such constraints in the phase plane, we mainly consider the results from \cite{zlajpah_implementation_1999}, from which we recall the most relevant developments. 

Acceleration constraints are given as
\begin{equation} \label{eq:acceleration_constraints}
    \lVert\ddot{\bm x}\rVert \le a_{max}.
\end{equation}
By inserting \eqref{eq:task_acceleration} in \eqref{eq:acceleration_constraints}, we find limits on $\ddot{\lambda}$:
\begin{equation} \label{eq:acceleration_bounds}
    \alpha_{x,min}(\lambda, \dot{\lambda}) \le \ddot{\lambda} \le \alpha_{x,max}(\lambda, \dot{\lambda}),
\end{equation}
where
\begin{equation}\label{eq:acceleration_bounds_old}
\begin{split}
    \alpha_{x,min}(\lambda, \dot{\lambda}) &= \frac{-\bm x'^T \bm x'' \dot{\lambda}^2} {\lVert\bm x'\rVert^2} + \\
    &- \frac{\sqrt{(\bm x'^T \bm x'')^2 \dot{\lambda}^4 + \lVert\bm x'\rVert^2 (a_{max}^2 - \lVert\bm x''\rVert^2 \dot{\lambda}^4)}}{\lVert\bm x'\rVert^2}, \\
    \alpha_{x,max}(\lambda, \dot{\lambda}) &= \frac{-\bm x'^T \bm x'' \dot{\lambda}^2} {\lVert\bm x'\rVert^2} + \\
    &+ \frac{\sqrt{(\bm x'^T \bm x'')^2 \dot{\lambda}^4 + \lVert\bm x'\rVert^2 (a_{max}^2 - \lVert\bm x''\rVert^2 \dot{\lambda}^4)}}{\lVert\bm x'\rVert^2}.
\end{split}
\end{equation}

By using \eqref{eq:unit_norm} and \eqref{eq:orthogonality}, the limits above become
\begin{align}\label{eq:acceleration_bounds_new}
    \alpha_{x,min}(\lambda, \dot{\lambda}) = - \sqrt{a_{max}^2 - \lVert\bm x''\rVert^2 \dot{\lambda}^4}, \\
    \alpha_{x,max}(\lambda, \dot{\lambda}) = + \sqrt{a_{max}^2 - \lVert\bm x''\rVert^2 \dot{\lambda}^4}.
\end{align}
Since the term under square root must be non-negative, the inequalities above also impose a limit on $\dot{\lambda}$:
\begin{equation} \label{eq:pseudovelocity_limit}
    \dot{\lambda} \le \sqrt{\nicefrac{a_{max}}{\lVert\bm x''\rVert}}.
\end{equation}

As far as task velocity constraints are concerned, they can be written as
\begin{equation}
    \lVert\dot{\bf x}\rVert \le v_{x,max},
\end{equation}
which, by using \eqref{eq:task_velocity} and \eqref{eq:unit_norm}, can be transformed into
\begin{equation} \label{eq:lambda_constraint}
    \dot{\lambda} \le v_{x,max}.
\end{equation}

Equations \eqref{eq:pseudovelocity_limit} and \eqref{eq:lambda_constraint} determine that, regardless of the path, pseudo-velocities cannot be arbitrarily chosen. Indeed, any PPT will be limited above by the so-called \emph{Maximum Velocity Curve (MVC)}
\begin{equation}
    \dot{\lambda}_{max}(\lambda) = \min\left({\sqrt{\frac{a_{max}}{\lVert{\bf x}''(\lambda)\rVert}}, v_{x,max}}\right),
\end{equation}
while its time derivative $\ddot{\lambda}$ will be subject to \eqref{eq:acceleration_bounds}. Therefore, TOTP is performed through the resolution of the following optimization problem:
\begin{equation} \label{eq:ocp}
\begin{split}
\min_{\ddot{\lambda} \in \mathcal{I}(\lambda, \dot{\lambda})} & \int_{0}^{L} \frac{1}{\dot{\lambda}(\lambda)}d\lambda, \\
\mbox{subject to} ~& 0 \leq \lambda \leq L, \\
& \dot{\lambda}(0) = \dot{\lambda}(L) = 0, \\
& 0 < \dot{\lambda} (\lambda) \leq \dot{\lambda}_{max} (\lambda), \\
& \mathcal{I}(\lambda, \dot{\lambda}) \triangleq \left[ \alpha_{x,min}(\lambda, \dot{\lambda}), \alpha_{x,max}(\lambda, \dot{\lambda}) \right].
\end{split}
\end{equation}
This problem can be demonstrated to have a bang-bang nature in the control $\ddot{\lambda}$ \cite{chen_general_1992} and can be solved with several techniques, being the switching points method \cite{slotine_improving_1989} and dynamic programming \cite{shinDynamicProgrammingApproach1986} the most used. The former is also the one we employ in this work.

\section{Index validation setup} \label{sec:index_validation_setup}

In virtue of Hypothesis \ref{hp:hl_comfort} and our definition of the human-likeness index in \eqref{eq:human-likeness}, we would like to answer the following questions.
\begin{itemize}
    \item Given two movements with significantly different values of $HL$, is the one with the largest value of $HL$ recognized as more comfortable than the other?
    \item Given two movements with similar $HL$ values, are they indistinguishable from the comfort standpoint?
\end{itemize}

For such validation, we recruited participants among the members of the Robotics Laboratory and students enrolled in various STEM and non-STEM curricula from the University of Salerno. All subjects agreed to participate voluntarily and provided formal consent by reading and signing a document outlining the study's purpose and the participants' rights\footnote{The experimental procedure had no potential to damage the mental or physical health of human participants. The research project did not involve vulnerable participants, nor did it have access to private or sensitive data for purposes that are not related to the project. The authors adhered to the code of conduct of the University of Salerno.}.
Moreover, participants were pre-selected based on the information provided in a preliminary questionnaire, accompanying this paper as supplementary material. This inquired about their motor skills and any potential upper body injuries that could have prevented them from experiencing the movements or providing an unbiased answer. Therefore, we excluded participants who had suffered from upper body injuries in the last 3 months. In the end, 76 subjects were enrolled.

The participants were divided in two separate groups. The first one, termed G1, consists of 8 subjects, 7 males and 1 female, aged between 23 and 33 years \textit{(mean 26.9, std $\pm$ 3.9)}.
The second one (G2), comprises 68 subjects, 33 males and 35 females, aged between 19 and 30 years \textit{(mean 22.4, std $\pm$ 4.2)}. Among them, 35 are enrolled in STEM and 33 in non-STEM curricula.
No subject was included in both G1 and G2.


\begin{figure}
\begin{center}
\includegraphics[width=\columnwidth]{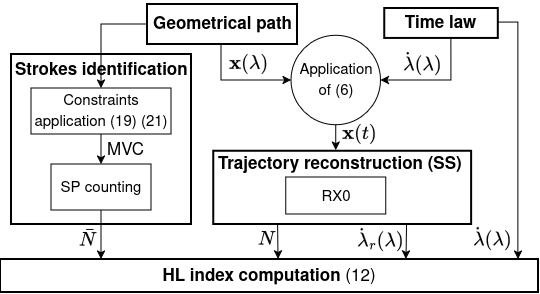}
\end{center}
\caption{$HL$ index computation process. The numbers in brackets refer to the equations used in the text.}\label{fig:hl_index_calculation}
\end{figure}






\begin{algorithm}
\SetAlgoLined
\LinesNumbered
\KwData{$\dot{\lambda}_h(\lambda), \bm{x}_h(\lambda)$}
$\dot{\lambda}_{UNI}(\lambda) \gets$ \eqref{eq:timestamps} \\
$\dot{\lambda}_{STOTPAC}(\lambda) \gets$ Integrate $\ddot{\lambda}$ from \eqref{eq:ocp} and scale \\
\ForEach{$\dot{\lambda}(\lambda) \in \lbrace \dot{\lambda}_h, \dot{\lambda}_{UNI}, \dot{\lambda}_{STOTPAC} \rbrace$}{
    $HL \gets$ Algorithm \ref{alg:hl_index} ($\dot{\lambda}(\lambda), \bm{x}_h(\lambda))$
}
\caption{Experimental dataset generation}
\label{alg:experimental_data_generation}
\end{algorithm}

\begin{algorithm}
\SetAlgoLined
\LinesNumbered
\KwData{$\dot{\lambda}(\lambda)$, $\bm{x}(\lambda)$}
$\bar{N} \gets$ Count strokes in $\bm{x}(\lambda)$ \\
$\lambda(t) \gets$ Compute from $\dot{\lambda}(\lambda)$, inverting \eqref{eq:time_law}\\
$\bm x(t) \gets \bm{x}\big(\lambda(t)\big)$ \\
$[\dot{\lambda}_{r}(\lambda), N] \gets$ Reconstruct $\bm x(t)$ using SS\\
$SNR \gets$ Compute SNR between $\dot{\lambda}(\lambda)$ and $\dot{\lambda}_r(\lambda)$ \\
$HL \gets $ Apply \eqref{eq:human-likeness} using $SNR$, $N$, and $\bar{N}$ \\
\caption{$HL$ index computation process}
\label{alg:hl_index}
\end{algorithm}

The methodology for the experimental dataset generation process is presented by the pseudocode in Algorithm \ref{alg:experimental_data_generation}, while the HL index computation process is shown in Fig. \ref{fig:hl_index_calculation} and accompanied by the pseudocode in Algorithm \ref{alg:hl_index}. G1 records human movements on a graphic tablet, as detailed next in Section \ref{sec:dataset_acquisition}. The geometrical features of the movements are then extracted and time-parametrized with both uniform and time-optimal velocity profiles, as described in Section \ref{sec:time_para_robotic_paths}. Thus, one human and two artificial/synthetic time laws are available for each geometrical path. Both human and artificial trajectories are then fed into the trajectory reconstruction software, which extracts the sigma-lognormal profiles needed to compute the $SNR$ in \eqref{eq:human-likeness}. The number of expected strokes $\bar{N}$ is identified from the geometrical path. Indeed, it corresponds to the number of target points because elementary strokes are the actions to reach them. In the stroke segmentation literature, target points are typically identified where significant changes in path curvature occur. Inspired by the functioning of the human vision system and the concept of saliency \cite{itti_1998}, algorithms have been proposed that can retrieve these segmentation points as the highest peaks in a curvature saliency map \cite{parziale_2022}. In our context, these target points, \ie $\bar{N}$, correspond to the switching points from maximum deceleration to maximum acceleration, identified on the MVC in the phase plane. The interested reader is referred to \cite{slotine_improving_1989} for more details on switching points identification. We note that this methodology does not require additional parameters to be estimated, and is agnostic regarding both trajectory reconstruction and stroke recognition algorithms. In the next two sections, we provide details for generating human and artificial trajectories by G1.

\subsection{Generation of human trajectories} \label{sec:dataset_acquisition}

Research on human movements typically requires two key components: the recording equipment and the specific motor task under examination. The selection of both components hinges on the application or aspect of the movement being investigated. For instance, motion trackers were used in a study on feeding manipulation tasks \cite{bhattacharjee_towards_2019}, while smart pads were utilized to diagnose Alzheimer's disease by analyzing drawing and handwriting movements \cite{cilia_diagnosing_2022}.

\begin{figure}[!t]
    \centering
  \subfloat[\label{fig:ll}]{%
       \includegraphics[width=0.32\linewidth,height=0.22\linewidth]{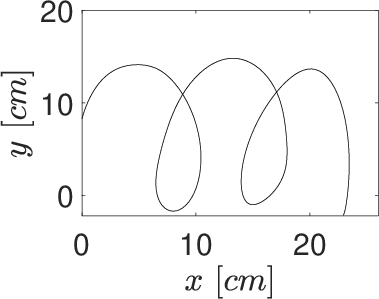}}
       \hspace{1px}
  \subfloat[\label{fig:5l}]{%
       \includegraphics[width=0.32\linewidth,height=0.22\linewidth]{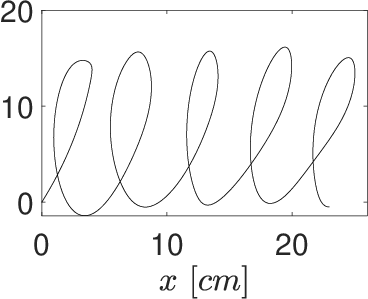}}
       \hspace{1px}
  \subfloat[\label{fig:eu}]{%
        \includegraphics[width=0.32\linewidth,height=0.22\linewidth]{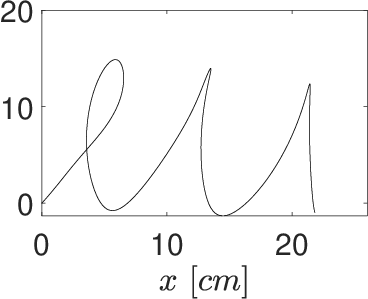}}
    \hfil
  \subfloat[\label{fig:ll_vel}]{%
        \includegraphics[width=0.32\linewidth,height=0.22\linewidth]{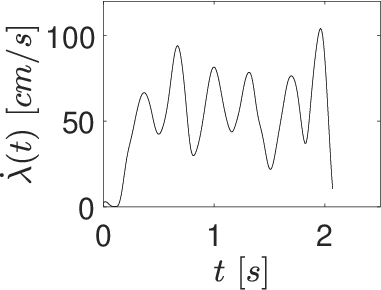}}
        \hspace{1px}
  \subfloat[\label{fig:5l_vel}]{%
        \includegraphics[width=0.32\linewidth,height=0.22\linewidth]{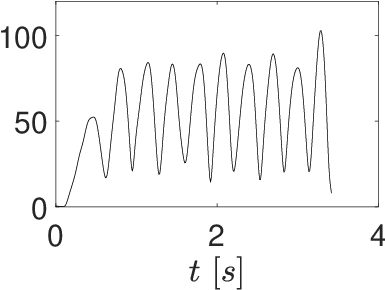}}
        \hspace{1px}
  \subfloat[\label{fig:eu_vel}]{%
        \includegraphics[width=0.32\linewidth,height=0.22\linewidth]{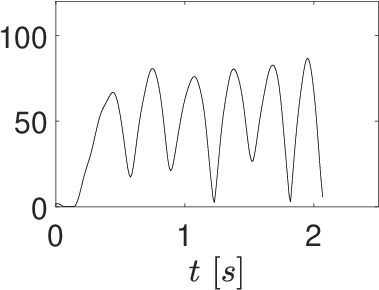}}
    \hfil
  \subfloat[\label{fig:ll_vel_uni}]{%
        \includegraphics[width=0.32\linewidth,height=0.22\linewidth]{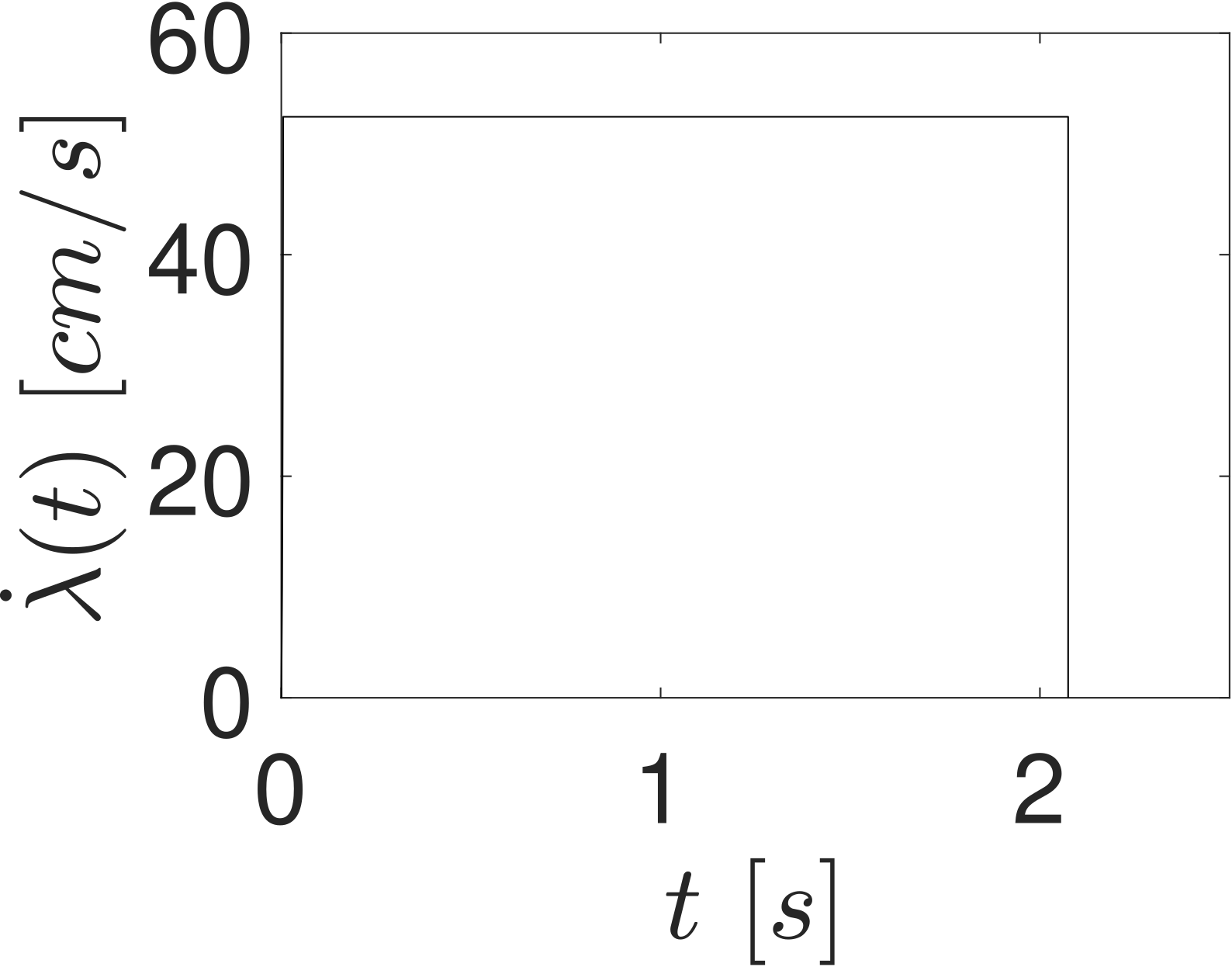}}
        \hspace{1px}
  \subfloat[\label{fig:5l_vel_uni}]{%
        \includegraphics[width=0.32\linewidth,height=0.22\linewidth]{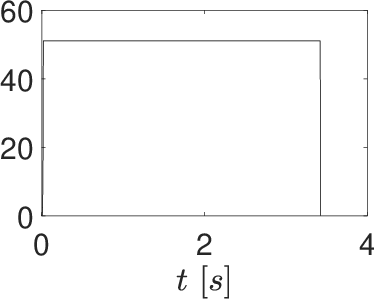}}
        \hspace{1px}
  \subfloat[\label{fig:eu_vel_uni}]{%
        \includegraphics[width=0.32\linewidth,height=0.22\linewidth]{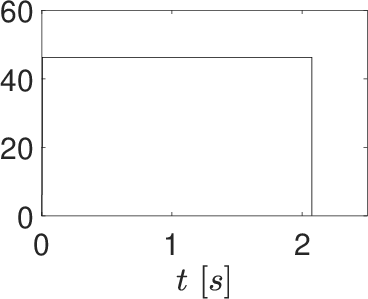}}
    \hfil
  \subfloat[\label{fig:ll_vel_B}]{%
        \includegraphics[width=0.32\linewidth,height=0.22\linewidth]{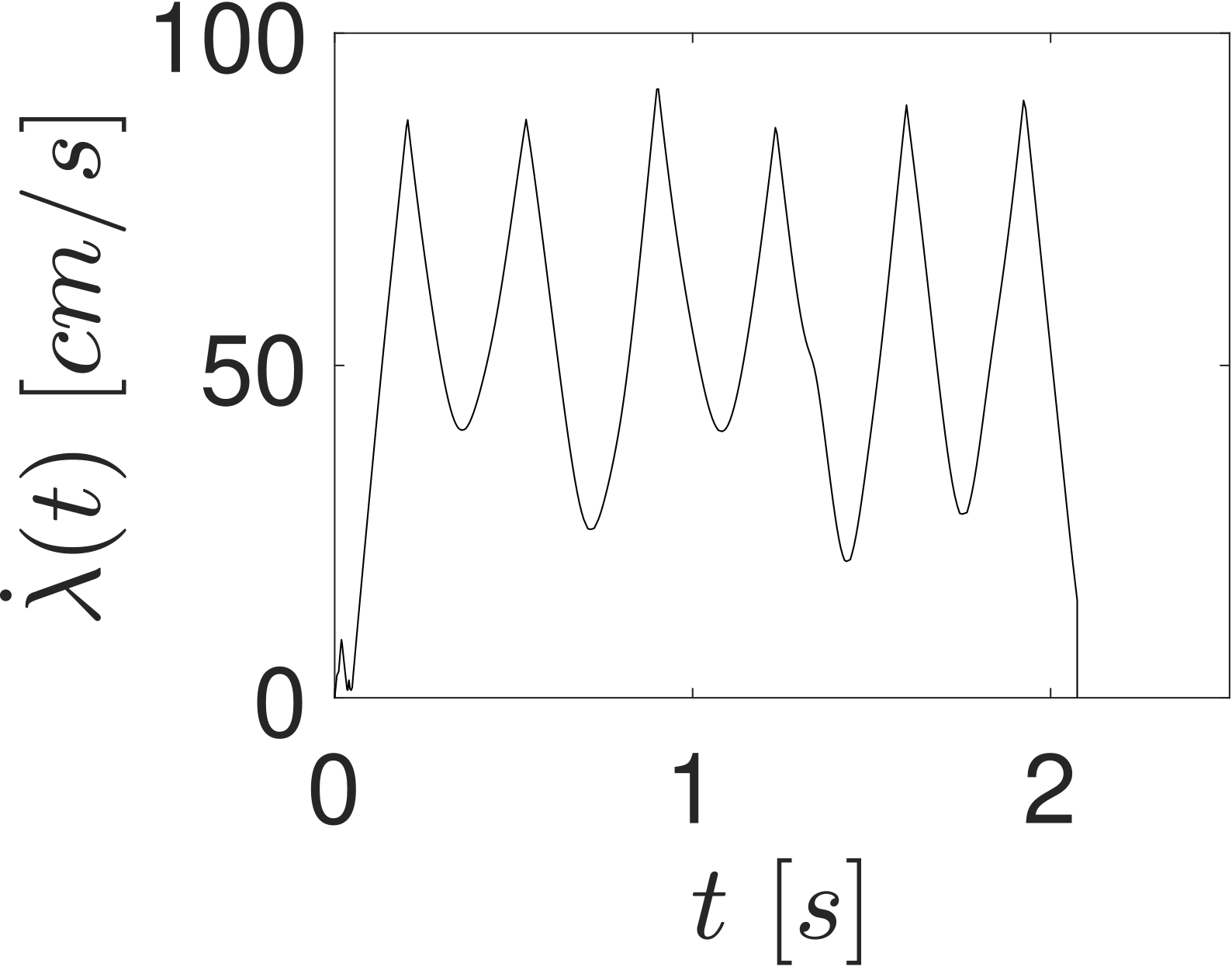}}
        \hspace{1px}
  \subfloat[\label{fig:5l_vel_B}]{%
       \includegraphics[width=0.32\linewidth,height=0.22\linewidth]{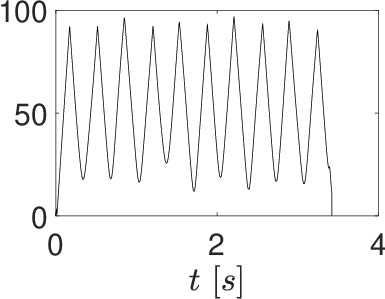}}
       \hspace{1px}
  \subfloat[\label{fig:eu_vel_B}]{%
        \includegraphics[width=0.32\linewidth,height=0.22\linewidth]{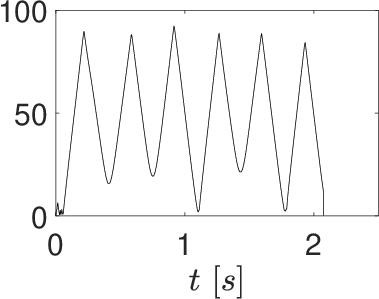}}
  \caption{(a)-(c) Human handwriting patterns with (d)-(f) corresponding tangential velocity, (g)-(i) Uniform tangential velocity and (j)-(l) STOTPAC tangential velocity, see Section \ref{sec:gen_artificial_traj}.}
    \label{fig:patterns}
\end{figure}

In our research, for the sake of practicality, we opt for a motor task involving drawing simple shapes ((a)-(c) in Fig.\ \ref{fig:patterns}).
Such shapes can be identified as a \emph{reversed spring} (RS), a sequence of \emph{5 connected ``l''} (5L), and the bigram \emph{eu} (EU). They vary by their respective levels of curvature within the pattern. For instance, the RS has mild curves, the EU has two points at infinite curvature, and the 5L exhibits, on average, an intermediate curvature compared to the other two.
Moreover, by selecting these shapes instead of handwritten words or goal-driven actions, both the subjects performing the tasks and those interacting with the robot can concentrate solely on the sensorimotor aspects of writing, avoiding cognitive overload, as previous research indicates that drawing objects with a semantic component activates cortical areas unrelated to motor execution but dedicated to the integration of semantic information \cite{harrington_neural_2009}. Notably, the selected shapes require precise synchronization of multiple elementary movements, making it challenging from a motor perspective. Moreover, choosing a 2D movement does not pose any limitations, as the kinematic theory of  rapid human movements has been shown to easily extend to 3D movements \cite{fischer_modeling_2020}.

Shapes were drawn using a ballpoint pen on a sheet of paper placed on an ink-and-paper WACOM Intuos 2 digitizing tablet, that has a sampling rate of 100 Hz. Motion data of the pen on the 2D tablet plane was recorded using MovAlyzeR v6.1 \cite{teulings_movalyzer_2019}. Each participant in G1 received instructions on writing the patterns shown in Fig.\ \ref{fig:patterns} (a)-(c), so to cover as much as possible of the A4 sheet surface. The participants were prompted by a computer beep for each writing instance, recording 10 repetitions per pattern, thus obtaining 80 samples of movements execution for each pattern. Post-processing removed the in-air movements at the beginning and end of each execution.

\subsection{Generation of artificial trajectories} \label{sec:gen_artificial_traj}

To generate the artificial handwriting trajectories corresponding to the selected human movements, we adopt the techniques introduced in Section \ref{sec:time_para_robotic_paths}, by taking all the human trajectories produced by G1 and replacing the original time law, so that the geometrical path does not change. To make TOTP time laws more similar to human ones, we select $a_{max}$ in \eqref{eq:acceleration_constraints} to be the maximum acceleration observed in the recorded human movement, and scale the time axis resulting from the resolution of \eqref{eq:ocp} to match the duration of the corresponding human time law. For the same reason, in case of the uniform velocity profile the value of $k$ in \eqref{eq:timestamps} is computed from the human movement duration, removing any bias associated with different durations when observing/experimenting with such trajectories.

In what follows, we refer to human-recorded trajectories as \emph{Human} and to the two artificial classes of handwriting trajectories as \emph{Uniform} and \emph{STOTPAC} (Scaled TOTP with Acceleration Constraints). In Fig.\ \ref{fig:patterns} (g)-(i), we report the tangential velocity profiles associated with the Uniform time law, while those associated with STOTPAC are in Fig.\ \ref{fig:patterns} (j)-(l).

\subsection{HL index evaluation} \label{sec:human_like_evaluation}

Tab.\ \ref{tab:SNR_numstrokes} displays the mean and standard deviation of the SNR in \eqref{eq:snr_reconstruction}, the number of strokes $N$ and the $HL$ index in \eqref{eq:human-likeness} for all Human trajectories executed by G1 and their artificial counterparts. The values are grouped by the patterns of Fig.\ \ref{fig:patterns} (a)-(c), for which the expected number of strokes in the motor program $\bar{N}$ (computed on the path) is also reported.


\begin{table}
\caption{Mean and standard deviation of SNR, number of strokes $N$ and $HL$ index for all human and artificial trajectories, grouped by pattern}
\label{tab:SNR_numstrokes}
\begin{center}
\setlength{\tabcolsep}{4pt} 
\begin{tabular}{c|c|c|c|c|c}
Method  & Pattern & $\bar{N}$ & $SNR$            & $N$              & $HL$             \\ \hline
Human   &         &           & 27.86 $\pm$ 2.08 & 12.28 $\pm$ 2.05 & 13.99 $\pm$ 2.54 \\
Uniform & RS      & 6         & 25.15 $\pm$ 1.71 & 18.34 $\pm$ 2.25 & 8.34 $\pm$ 1.10  \\
STOTPAC &         &           & 26.42 $\pm$ 1.04 & 13.34 $\pm$ 1.48 & 12.03 $\pm$ 1.45 \\ \hline
Human   &         &           & 27.16 $\pm$ 1.84 & 18.24 $\pm$ 2.57 & 15.17 $\pm$ 2.23 \\
Uniform & 5L      & 10        & 23.26 $\pm$ 1.66 & 32.78 $\pm$ 3.26 & 7.18 $\pm$ 0.94  \\
STOTPAC &         &           & 24.90 $\pm$ 1.76 & 19.35 $\pm$ 2.88 & 13.10 $\pm$ 1.85 \\ \hline
Human   &         &           & 26.17 $\pm$ 1.99 & 12.16 $\pm$ 2.30 & 13.37 $\pm$ 2.75 \\
Uniform & EU      & 6         & 21.12 $\pm$ 1.61 & 21.10 $\pm$ 2.53 & 6.07 $\pm$ 0.70  \\
STOTPAC &         &           & 24.89 $\pm$ 1.65 & 13.40 $\pm$ 1.93 & 11.35 $\pm$ 1.66 \\ \hline
\end{tabular}
\end{center}
\end{table}

\begin{figure*}
    \centering
  \subfloat{%
       \includegraphics[width=0.3\linewidth]{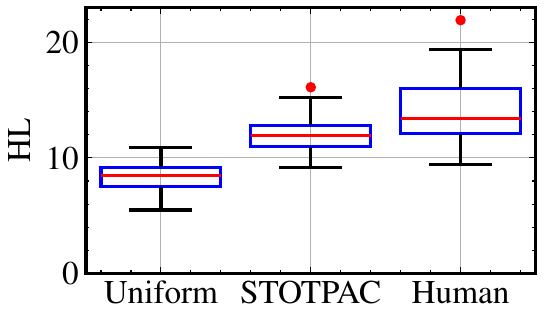}}
    \hfill
  \subfloat{%
       \includegraphics[width=0.3\linewidth]{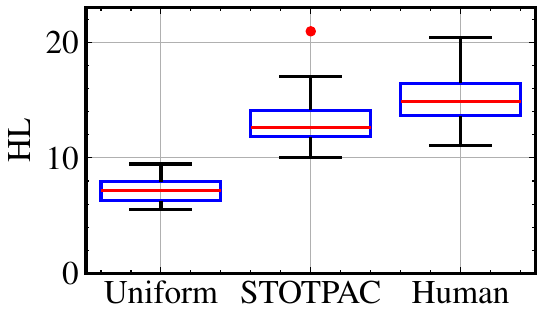}}
    \hfill
  \subfloat{%
       \includegraphics[width=0.3\linewidth]{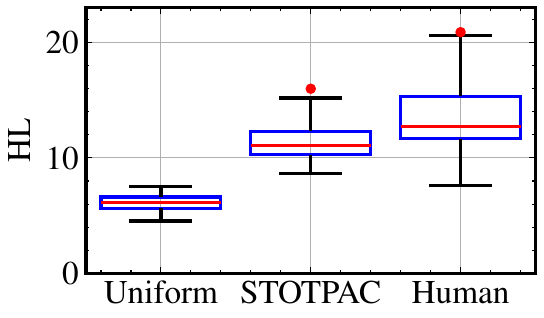}}
  \caption{Boxplot of $HL$ index in \eqref{eq:human-likeness} calculated for patterns in Fig.\ \ref{fig:ll}, \ref{fig:5l} and \ref{fig:eu}, ordered from left to right, respectively.} 
    \label{fig:hl_boxplot}
\end{figure*}

The same data are shown in Fig.\ \ref{fig:hl_boxplot}, in which the distributions of the $HL$ index are reported for each pattern and class of time laws in the form of box plots.


Notably, since the STOTPAC trajectories have been designed to resemble human ones, and because they naturally take into account the path curvature, we observe a high degree of similarity in the $HL$ index between Human and STOTPAC trajectories. This suggests that the latter may be difficult to distinguish from Human ones, making them more suitable for approximating human time laws.

Proposition \ref{prop:max_hl_implies_human_likeness} suggests that if a planner can generate sigma-lognormal velocity profiles with a limited amount of strokes, ideally the number of strokes that would be comprised in a fully-developed motor program, then the resulting movement will be human-like. We thus pose the question: if a movement is designed to maximize the $HL$ index, is the movement perceived as comfortable by humans? Hence, we posit

\begin{proposition} \label{prop:hl_index_comfort_index}
If, for a given movement, the $HL$ index is maximized, the movement is comfortable.
\end{proposition}

In virtue of Hypothesis \ref{hp:hl_comfort} and Proposition \ref{prop:max_hl_implies_human_likeness}, we expect Proposition \ref{prop:hl_index_comfort_index} to hold. Through the design of suitable experiments, we aim to provide empirical evidence.

\subsection{Interaction experiment} 
\label{sec:interaction}

With the focus on robot acceptance in collaborative tasks, we structure our experiments with a robot that passively moves the arm of the involved subjects (to activate their proprioceptive receptors), asking them about their perceived comfort. 


The experiments are designed as comparison tests, in which human subjects are presented with two handwriting motions and asked to select the most comfortable one according to their self-evaluation \cite{wangLearningComfortHuman2019}.
Comparison tests are preferred because humans struggle to assess movements when they lack a reference \cite{mignoneObservationVsInteraction2023a}. Moreover, formulating the question in terms of comfort, instead of similarity to a human, allows for the subject to express a bias-free preference.

\begin{figure}
\begin{center}
\includegraphics[width=\columnwidth]{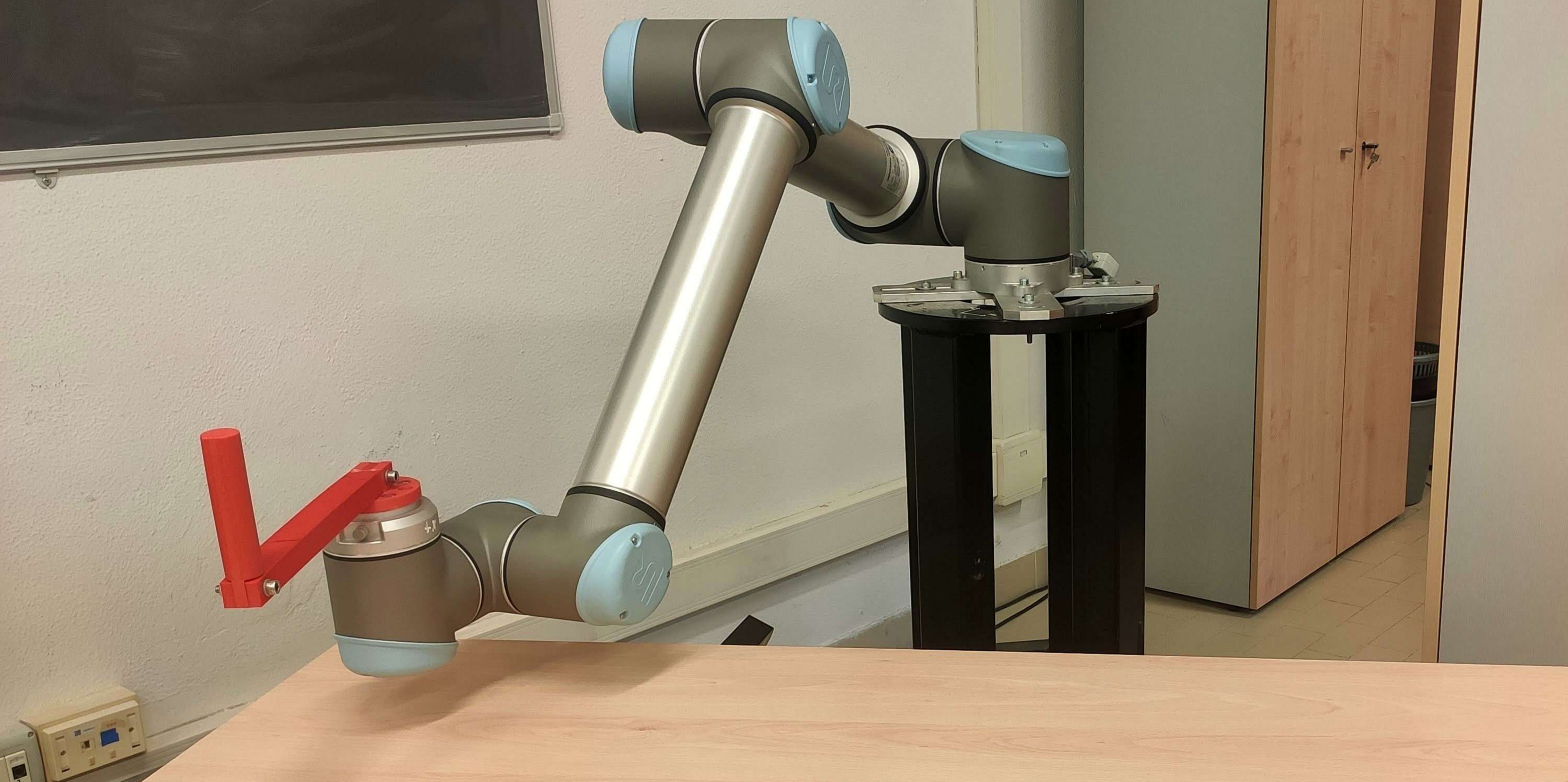}
\end{center}
\caption{Human-Robot interaction experiment setup.}\label{fig:interaction_experiment_setup}
\end{figure}

The G2 group is involved in the experimental procedure, together with a velocity-controlled 6-DOF UR10 robot, which is supplied with both Human and artificial trajectories. During the experiment, subjects are required to actively engage with the robot, maintaining contact with its end-effector,  which is custom-designed and 3D-printed, during the execution of movements. Since an underlying assumption of this work is that the human behavior is encoded in the velocity information \cite{plamondon_generation_1998}, we configure the robot controller to perform velocity control, that is to control velocity directly. The robot executes the handwriting trajectories within its free space, without interacting with any surrounding objects, on a virtual horizontal plane. To ensure safety, a physical barrier is placed between the robot and the human, preventing the participant from entering the robot's workspace with their entire body. Additionally, the robot is programmed to operate near its workspace boundary to limit its extension towards the human. The end-effector is stiff to avoid the introduction of artificial dynamics, yet it maintains a degree of fragility to facilitate breakage in the event of a collision, preserving the participant's safety. The experiment setup is as in Fig.\ \ref{fig:interaction_experiment_setup}.

\section{Experimental Setup} \label{sec:experimental_setup}

\begin{figure}
    \centering
  \subfloat{%
       \includegraphics[width=1\linewidth]{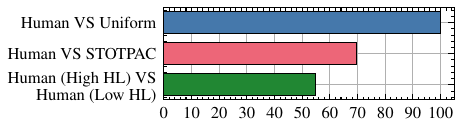}}
    \\[-2pt]
  \subfloat{%
       \includegraphics[width=1\linewidth]{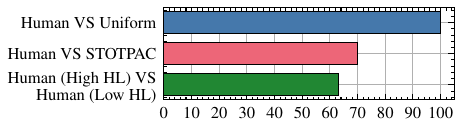}}
    \\[-2pt]
  \subfloat{%
       \includegraphics[width=1\linewidth]{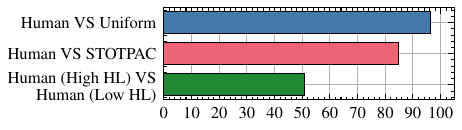}}
    \\[-5pt]
  \caption{Preference rate for the most comfortable trajectory of a pair in all campaigns, evaluated on the patterns in Fig.\ \ref{fig:ll}, \ref{fig:5l} and \ref{fig:eu}, ordered from top to bottom, respectively.} 
    \label{fig:results}
\end{figure}

Three campaigns of experiments have been conducted:
\begin{enumerate}
    \item \emph{Human-Uniform}, involves the robot executing 30 trajectories, in the form of 15 pairs (5 per pattern), each composed of a Human and its Uniform counterpart.
    \item \emph{Human-STOTPAC}, involves the robot executing 30 trajectories, in the form of 15 pairs (5 per pattern), each composed of a Human and its STOTPAC counterpart.
    \item \emph{Human-Human}, involves the robot executing 36 trajectories, in the form of 18 pairs (6 per pattern), each composed of a Human with higher $HL$ and a Human with lower $HL$.
\end{enumerate}

We select the trajectories shown during the experimental campaigns from the G1 dataset, collected as in Section \ref{sec:dataset_acquisition}. This selection is carried out to maximize the subjects in G1 and patterns coverage, while exploring the whole $HL$ range, as presented in Section \ref{sec:human_like_evaluation}. In particular, we classify pairs based on three levels of $HL$ difference: low, medium, and high, defined relatively to the pairs in the same campaign. Accordingly, low-to-medium and medium-to-high thresholds are $(5,10)$, $(1.5,5)$, and $(0.5,5)$ for the first, the second, and the third campaign, respectively. In the context of the third campaign, we also address instances where a comparison is drawn between two human trajectories, either executed by the same subject or by two distinct ones. This allows us to account for 3 (number of patterns) $\times$ 3 (levels of difference of $HL$) $\times$ 2 (case in which the subject is the same or not) = 18 trajectories.


Each participant completes all three campaigns, thus accounting for a total of 48 queries, i.e., single comparisons between two trajectories. 
The order of execution of the two trajectories in a query and of queries in a campaign is randomized among participants, to avoid ordering biases.
In each query, participants are asked: ``In which of the two movements did you feel more comfortable in interacting with the robot?''. 

Participants enter the experimental room individually to ensure that their decision-making is not influenced by the opinions of other participants. To avoid potential fatigue and attentional drop-off effects, that could affect the results, a single pause is included within each campaign, after the first half of queries (8 for \emph{Human-Uniform} and \emph{Human-STOTPAC}, 9 for \emph{Human-Human}). Furthermore, each campaign is conducted one week apart to minimize carryover effects.

Before each campaign, the operator presents the protocol to the participant and shows the three patterns, as seen in Fig.\ \ref{fig:patterns} (a)-(c). The patterns, printed in their original size, are placed on the desk below and parallel to the execution plane as a visual reference. This approach enables the participants to familiarize themselves with the geometrical features of the trajectories, so preventing the participants from creating a mental representation of the movements while they are experiencing it, a phenomenon that could potentially interfere with their decision-making process or result in reduced attentiveness. 

The protocol adopted across all the experimental campaigns is presented in written form to the subject before performing each campaign: \emph{
``The robot will perform the 3 writing patterns shown herein as if holding a pen and writing on a paper sheet lying on the desk.
Each pattern will be executed twice, and throughout each robot execution, you shall hold the orange handle as if it were a pen. Grasp it firmly to prevent it from slipping out of your hand. After the second execution terminates, you can let go off the grip. 
You will then be asked to answer the following question: ``In which of the two movements did you feel most comfortable interacting with the robot?'' If you cannot provide a definitive answer, you can ask the robot to execute the pattern again.
The robot has been positioned so that it cannot touch you, and, for safety reasons, there will be an operator present at all times, ready to stop the robot if there are any anomalies.
When you feel ready, we can start. The first experiment will begin in a few seconds.''}



\section{Results Analysis} \label{sec:results_analysis}

\subsection{Consistency between HL index and user preferences}

The outcomes of the experiments, in terms of preference rate for the most comfortable trajectory in a pair, for the three campaigns, are summarized in Fig.\ \ref{fig:results}. Except for one outlier associated with the EU pattern, i.e.\ Fig.\ \ref{fig:results} (bottom), our trials show a consistent preference for the Human trajectories in the Human-Uniform comparisons. When replacing the Uniform profile with the STOTPAC one, the participants still globally prefer the Human trajectories, but, in several trials, they feel STOTPAC trajectories more comfortable than Human ones. Lastly, when comparing two Human trajectories, as expected, subjects seem to have not a clear preference between one trajectory or the other, but a slight preference for the Human trajectory with higher $HL$. We observe that this trend is globally consistent with the distributions of the $HL$ values reported in Fig.\ \ref{fig:hl_boxplot}, i.e., Human trajectories exhibiting, on average, a larger $HL$ value, are perceived as more comfortable by the general population, and, compared to Uniform ones, STOTPAC trajectories, presenting an intermediate average $HL$ value, are more frequently ``confused'' with Human ones.

\subsection{Comparison with jerk}

Since mean absolute jerk (MAJ) is a common a-priori metric to assess the comfort of movements, we conduct a dedicated analysis to compare it with both the proposed HL index and the outcome of interaction experiments. Therefore, we focus on the 15 trajectories experimented by volunteers in the interaction experiments of the \emph{Human-Uniform} and \emph{Human-STOTPAC} campaigns. Let $j_U$ and $j_S$ be the MAJ for Uniform and STOTPAC trajectories, respectively. While there is a general agreement between HL index, MAJ and user preference, there might be cases in which trajectories with higher jerk are preferred more frequently by users than trajectories with lower jerk. Tab.\ \ref{tab:jerk_comparison} reports two of such cases from our dataset, i.e., STOTPAC trajectories are preferred more frequently than Uniform trajectories ($18.42\%$-$28.95\%$ vs.\ 0\%, respectively) in comparisons with Human trajectories, despite being $j_S > j_U$\footnote{In the case of Uniform trajectories, the MAJ is computed on the executed trajectory, including the short segments necessary to reach the peak velocity and to return back to zero. Only these segments contribute to the jerk computation, as it is otherwise zero.}. Conversely, the HL index, by capturing more fundamental aspects of the human movement, better aligns with the user preference.

\begin{table}
\centering
\caption{Comparison between HL, MAJ and user preference rate}
\label{tab:jerk_comparison}
\setlength{\tabcolsep}{2pt}
\begin{tabular}{c|cccc|c|c}
 & & & & & Preference rate & Preference rate \\ 
ID & & Human & STOTPAC & Uniform & for Human & for Human  \\
 & & & & & wrt STOTPAC & wrt Uniform \\
 \hline
01WL & HL & 16.57 & 14.72 & 8.05 & 81.58\% & 100\% \\
 & MAJ & 31.35 & 46.72 & 43.75 & & \\
\hline
01TL & HL & 15.63 & 13.90 & 6.31 & 71.05\% & 100\% \\ 
 & MAJ & 62.80 & 90.41 & 77.81 & &\\ 
\end{tabular}
\end{table}

\subsection{Impact of path curvature on HL index}

We note that the preference for the Human trajectory in the Human-STOTPAC comparison campaign showcases some variability across patterns, and, specifically, the subjects feel the STOTPAC time laws particularly uncomfortable when applied to the EU pattern. 
To explain this different behavior in the EU pattern case, we rely on unsolicited comments that some participants made when answering the Human-STOTPAC queries related to such a pattern. In particular, without knowing the agent that generated the trajectories, they reported that STOTPAC ones felt more jerky, which was influencing their perception of comfort and, hence, guiding their preference towards Human ones. Indeed, because of its bang-bang nature, TOTP accentuates acceleration and deceleration as the path curves get tighter. Among the patterns, the EU one presents the tightest curves at the two angular points delineating the upper edges of the `U' letter. Whereas a human would naturally approach such points with deliberate, controlled deceleration and acceleration, TOTP produces abrupt velocity variations, that are likely the cause of the jerkiness reported by subjects. This detailed aspect is only weakly captured by the proposed $HL$ index in \eqref{eq:human-likeness} (see Tab.\ \ref{tab:SNR_numstrokes}), while it is a core factor that subjects may consider when evaluating comfort.

\subsection{HL index as comfort predictor}

We want to assess the performance of the proposed $HL$ index in indicating comfort. In particular, said $HL_h$ and $HL_l$ the $HL$ values for the two trajectories in the comparison, and $\Delta HL = HL_h - HL_l > 0$, we assess whether the binary variable ``preference for the trajectory with $HL_h$'' ($p_h$) correlates with the continuous variable $\Delta HL$, i.e.\ humans are more likely to perceive the most human-like trajectory as comfortable as the distance in terms of $HL$ increases. On the other hand, if the two trajectories are similar in terms of $HL$, they tend to be ``confused'', i.e.\ humans do not exhibit a clear preference for one trajectory or the other. In other words, we want to assess whether $\Delta HL$ is a predictor for $p_h$. The results of the three campaigns are grouped by pattern, and then subject to confirmatory analysis by a logistic regression. The dependent variable $p_h$ is equal to 1 if the preferred trajectory is the one with $HL_h$, 0 otherwise. The logistic regression results, as shown in Tab.\ \ref{tab:logistic_regression_coefficients}, indicate that the probability of feeling the trajectory with $HL_h$ more comfortable has a statistically significant association with $\Delta HL$. This is confirmed by a deviance test, which determines whether our models fit significantly better than constant ones (i.e.\ no correlation condition). The deviance test is performed on each pattern, with $p = 1.05 \cdot 10^{-16}$ for the RS one, $p = 4.14 \cdot 10^{-17}$ for the 5L one and $p = 4.71 \cdot 10^{-8}$ for the EU one. These p-values indicate that each model significantly differs from a constant, thus proving that $\Delta HL$ is a predictor for $p_h$.
Odds Ratio (OR) and 95\% Confidence Interval of OR (95\% CI OR), for each model, are reported in Tab.\ \ref{tab:logistic_regression_coefficients}. We use 4 decimal places to keep statistical precision. 

A post-hoc power analysis is conducted using G*Power version 3.1 to evaluate the achieved statistical power of the logistic regression analysis performed for each pattern \cite{faul2009statistical}. Each analysis is based on the total sample size, i.e., 1088, and a significance level $\alpha = 0.05$. The odds ratio, which is the observed effect size, is equal to $1.2120$, $1.1721$, and $1.1005$ for RS, 5L, and EU patterns, respectively. The achieved power $(1-\beta)$ is calculated to be 100\%, 100\%, and 99.9\% for RS, 5L, and EU patterns, respectively. This suggests that the three logistic regression tests have sufficient power to detect the observed effect.

\begin{table}
\centering
\caption{Logistic regression results for RS, 5L and EU patterns. \\ Dependent variable = $p_h$}
\label{tab:logistic_regression_coefficients}
\begin{tabular}{c|c|c|c}
 & Predictor& OR & 95\% CI OR \\ 
 \hline
RS & $\Delta HL$ & 1.2120 & {[}1.1539, 1.2732{]} \\ 
5L & $\Delta HL$ & 1.1721 & {[}1.1253, 1.2209{]} \\ 
EU & $\Delta HL$ & 1.1005 & {[}1.0623, 1.1401{]} \\ 
\end{tabular}
\end{table}

\subsection{HL index sensitivity}

An additional analysis is performed on the Human-Human comparisons campaign. The aim is to assess whether marked differences in terms of the $HL$ index between two trajectories in a comparison correspond to differences in the perceived comfort. In other words, we aim to assess whether a threshold of $\Delta HL$ exists at which two trajectories are clearly perceived as different from a comfort perspective.

Since we do not observe marked differences in the comfort perception between the two patterns, for each single trajectory pair, let $P_h$ be the rate of preference for the trajectory with $HL_h$ across subjects, regardless of the pattern, i.e.\
\begin{equation}
    P_h = \frac{\sum_{i=1}^{\left| \mbox{G2} \right|} p_{h,i}}{\left| \mbox{G2} \right|},
\end{equation}
where $\left| \mbox{G2} \right|$ is the cardinality of G2, and $p_{h,i}$ is the $i$-th subject preference for the trajectory with $HL_h$. We group all trajectory pairs into two groups, having either $\Delta HL \leq T$ or $\Delta HL > T$, where $T=0.600$ is an empirically selected threshold that well separates the classes. We find that 6 pairs fall into the former group, while 12 pairs fall into the latter. For each group, we compute the average of the rates $P_h$, i.e.\ $\mu_l = 40.50$ and $\mu_h = 63.92$, respectively. After assessing normality through a Shapiro-Wilk test across both the whole sample and within every single class, identified by the threshold, we conduct a two-tailed \emph{t-Student test} for the difference between the population means $\mu_l$ and $\mu_h$, with a significance level $\alpha=0.05$. The null hypothesis $H_0: \mu_l - \mu_h = 0$ is rejected with a p-value of $0.0119$, proving that there is statistical significance that the two groups' means are different. In line with the commented results of the logistic regression test, we confirm that two human trajectories can be distinguished in terms of comfort if they are ``different enough'' in terms of the proposed $HL$ index.


\subsection{Accuracy of HL index in single comparisons}

While a global correlation between the proposed $HL$ index and comfort is observed, as demonstrated by our analyses, the former is not guaranteed to be an index of comfort in single comparisons, mainly because the perception of comfort is user-dependent. Indeed, we observed subjects preferring the STOTPAC trajectory even in cases where its $HL$ index was remarkably lower than its Human counterpart. A clear example of this is found in subject 28 from the Human-STOTPAC campaign. In query 12, the subject preferred STOTPAC (HL: 13.50) over Human (HL: 22.20). This limitation can also be explained in light of Remark \ref{rem:sufficient_condition}: some users might ``feel'' features of the experienced movement that make it comfortable, even though its $HL$ index is low. However, our results show that the probability of such an event tends to zero as the distance in $HL$ between the two trajectories is maximized, thus confirming Proposition \ref{prop:hl_index_comfort_index}.

\subsection{Summary}

We designed our $HL$ index to be an index of similarity to a motor program, hence, from Proposition \ref{prop:hl_index_comfort_index}, it follows that if the similarity to a motor program is maximized, the movement is perceived as comfortable. In turn, these results also empirically validate Hypothesis \ref{hp:hl_comfort}. Ultimately, the issue of acceptance is resolved through the similarity to a motor program.



\section{Conclusion and future work} \label{sec:conclusion}

The issue of the acceptance of robots by human operators is a multifaceted one that is strongly linked to how humans perceive motion. Additionally, the level of comfort associated with a given movement, to which acceptance directly links, highly depends upon an individual's unique perception across a multitude of distinct factors. These properties are challenging to quantify, and are thus expected to require examples from humans. Based upon the kinematic theory of human movements, we proposed an index whose maximization provides a sufficient condition for a movement to be comfortable, without the need for humans to have an experience of it. We solved the problem of comfort by measuring the similarity of a movement to a mental motor program. In addition, our approach associates comfort with the movement's time law, which enables any movement to become comfortable, regardless of the movement's geometry. To validate our index, we conducted experiments with human subjects, who were instructed to interact with a robot and to evaluate comfort over human-recorded and artificially-planned trajectories.

Despite its limitation in capturing user-dependent factors, our results show that our index is indeed an index of comfort. In particular, according to it, human-recorded trajectories are classified as more comfortable than artificial ones, and this is confirmed by the direct experience of our sample of users. We specify that our results cannot be generalized to subjects whose motor program is not developed enough (\eg young children) or whose execution is affected by physical disabilities or stiff muscles (\eg elderly or disabled individuals). 

Moreover, we recognize that different user-dependent factors exist, such as different individuals' movement habits, physiological disparities, and psychological state fluctuations, that could influence the subjects' perception of movement. These factors may significantly influence comfort perception and the effectiveness of the HL index. Although these factors were not examined in our study, we laid the foundation for further investigations by collecting relevant information through the administered questionnaire, aimed at assessing some of these factors and evaluating participants’ motor competence.

With a focus on future developments on the topic of robot acceptance in HRI, the index's capability to provide an a priori characterization of trajectories makes it fit to be employed as a performance index for human-like trajectories' generation algorithms, providing an objective evaluation of such methods and facilitating their comparison. Such an index could be particularly valuable in motion planning for exoskeletons, especially in rehabilitation contexts, where therapies might require tracking a prescribed path for functional rehabilitation. As the HL index is also correlated to metabolic energy expenditure, future work might investigate how the index, applied to robotic trajectories, compares to direct measures of power consumption.
 
The potential applications of this index extend beyond robotic ones, an example being the generation of synthetic handwriting. By training generative networks to generate samples whose index is above a given threshold, the likelihood of generating implausible samples is reduced, thereby avoiding the introduction of artificial variability in the training data. Moreover, our proposed human likeness evaluation system in its entirety could be deployed to medical centers. When integrated with a motion capture system, it can be used to quantify arm motion pathologies that result in kinematic deviations from the normal range of motion observed in healthy upper limbs.

Our experiments also showcase that not all artificially-generated trajectories are equivalent in terms of comfort and that time-optimal parametrizations are sometimes good candidates to replace human movements while preserving comfort. To this end, it is pertinent to investigate new trajectory planning algorithms that would target the maximization of our index, possibly overcoming the limitations associated with time-optimal movements.


The fact that we can only define a sufficient condition for a movement to be comfortable makes our index more suitable for synthesis than analysis of movements. This prompts further inquiry into movement features that the similarity with the motor program does not address, but that are likewise important for comfort. This will be the subject of future research.

\section*{Acknowledgments}

The authors would like to thank Giovanni Mignone for conducting preliminary experiments that helped us lay the foundations of this study, and Gianluca Pallante for engaging in fruitful discussions on our statistical analyses.



\IEEEpubidadjcol






\bibliographystyle{IEEEtran}
\bibliography{IEEEabrv, OtherAbbrv, bibliography}

\end{document}